%% file: main.tex
\documentclass[10pt,conference,letterpaper]{IEEEtran}
\IEEEoverridecommandlockouts
\usepackage{cite}
\usepackage{amsmath,amssymb,amsfonts}
\usepackage{algorithmic}
\usepackage{graphicx}
\usepackage{textcomp}
\usepackage{subfig}
\usepackage{xcolor}
\usepackage{svg}
\def\BibTeX{{\rm B\kern-.05em{\sc i\kern-.025em b}\kern-.08em
    T\kern-.1667em\lower.7ex\hbox{E}\kern-.125emX}}

\usepackage[hyphens]{url}
\usepackage{array}
\usepackage{xspace}
\usepackage{balance}
\usepackage[absolute,overlay]{textpos}

\newcommand{\ie}{{\mbox{i.e., }\xspace}}
\newcommand{\eg}{\mbox{e.g., }\xspace}

\newcommand{\TOOL}{\mbox{RL-IoT}\xspace}

\begin{document}

\TPshowboxestrue
\TPMargin{0.3cm}
\begin{textblock*}{15.5cm}(3cm,1.4cm)
\footnotesize
\bf
\definecolor{myRed}{rgb}{0.55,0,0}
\color{myRed}
\noindent
Please cite the published version of this article at: 
Giulia Milan, Luca Vassio, Idilio Drago, Marco Mellia. RL-IoT: Reinforcement Learning to Interact with IoT Devices. 2021 IEEE International Conference on Omni-Layer Intelligent Systems (COINS), 2021, DOI: \url{https://doi.org/10.1109/COINS51742.2021.9524260}
\end{textblock*}

\title{\vspace{1.5cm}
RL-IoT: \\
Reinforcement Learning to Interact with IoT Devices}

\author{\IEEEauthorblockN{Giulia Milan}
\IEEEauthorblockA{\textit{Politecnico di Torino} \\
Turin, Italy \\
giulia.milan@polito.it}
\and
\IEEEauthorblockN{Luca Vassio}
\IEEEauthorblockA{\textit{Politecnico di Torino} \\
Turin, Italy \\
luca.vassio@polito.it}
\and
\IEEEauthorblockN{Idilio Drago}
\IEEEauthorblockA{\textit{University of Turin} \\ 
Turin, Italy \\
idilio.drago@unito.it}
\and
\IEEEauthorblockN{Marco Mellia}
\IEEEauthorblockA{\textit{Politecnico di Torino} \\
Turin, Italy \\
marco.mellia@polito.it}
}

\maketitle

\input{01_abstract.tex}


\input{02_introduction}
\input{03_related}
\input{04_methodology}
\input{05_results}
\input{06_conclusions}

\bibliographystyle{ieeetr}
\bibliography{main}

\end{document}

%% file: 01_abstract.tex
\begin{abstract}

Our life is getting filled by Internet of Things (IoT) devices. These devices often rely on closed or poorly documented protocols, with unknown formats and semantics.
Learning how to interact with such devices in an autonomous manner is the key for interoperability and automatic verification of their capabilities. In this paper, we propose \TOOL, a system that explores how to automatically interact with possibly unknown IoT devices. We leverage reinforcement learning (RL) to recover the semantics of protocol messages and to take control of the device to reach a given goal, while minimizing the number of interactions. 
We assume to know only a database of possible IoT protocol messages, whose semantics are however unknown. \TOOL exchanges messages with the target IoT device, learning those commands that are useful to reach the given goal.
Our results show that \TOOL is able to solve both simple and complex tasks. With properly tuned parameters, \TOOL learns how to perform actions with the target device, a Yeelight smart bulb in our case study, completing non-trivial patterns with as few as 400 interactions. \TOOL paves the road for automatic interactions with poorly documented IoT protocols, thus enabling interoperable systems.
\end{abstract}

%% file: 02_introduction.tex
\section{Introduction}

The popularity of IoT devices keeps growing at a fast pace, with the number of connected devices projected to be around 31 billion units worldwide by 2025. IoT devices are present in many IT systems, from smart homes to drones, from industry 4.0 scenarios to medical systems.\footnote{\url{https://www.statista.com/statistics/1101442/iot-number-of-connected-devices-worldwide}}
These devices rely on multiple standard protocols and technologies \cite{IoTlayers}, such as MQTT, CoAP and XMPP, but often they implement proprietary and not well-documented protocols whose semantics may be obscure.

A general approach for learning how to interact with IoT devices would represent an important step for many applications, including interoperability and cybersecurity. In the literature, this problem lies under the umbrella of protocol reverse engineering, i.e., the process of learning the protocol used by an application, having no or limited access to the protocol specification~\cite{ProtocolFieldInference,Polyglot,AutomaticContextAware}. 
For interoperability purposes, one often faces a simplified version of the problem, in which \emph{some} information about the protocol is indeed available. For instance, protocol messages and syntax may be public, but with little information about {\it protocol semantics}. Equally, even if some protocol information may be available, finding the precise operations  providing a particular functionality may be a hard task due to poor documentation.  

In this work, we build a system capable of learning by experience how to interact with IoT devices.
In details, given i) a target IoT device, e.g., a smart bulb, ii) a superset of protocol messages (not all of them supported by the target device), iii) a communication network, and iv) a feedback channel, we want to learn the specific sequence of messages that allows us to change the IoT device settings according to a desired sequence of states.
At the end, the system shall learn these messages in the shortest possible time, ultimately unveiling the semantics of each message.

To reach our goal, we rely on reinforcement learning (RL)~\cite{SuttonBarto}. A \emph{learner} stimulates the device and observes how it reacts, obtaining a positive (negative) reward when the device does (does not) perform the desired action.
We assume to receive a feedback from the device, for instance having a side channel to observe how its status changes (e.g., a camera looking at the smart bulb) or a feedback channel directly offered by the IoT protocol. 
More formally, RL builds an internal state-machine representing a portion of the IoT protocol.
The learner's goal is to discover how to navigate the state-machine, finding the best (e.g., shortest) sequence of actions to reach our goal.

We present \TOOL, a RL-based framework to automatically interact with IoT devices. We focus on a case study of a Yeelight smart bulb, which offers a proprietary protocol, generically documented for all Yeelight devices.
We present the design of \TOOL and offer a thorough set of experiments, comparing different RL methods, tuning parameters, and showing that \TOOL is effective to control the smart bulb, successfully completing both simple and complicated sequences of actions. 

Results show that not only \TOOL is able to find the optimal sequence of commands to control the device, but also discover multiple solutions, combining commands that at a first sight are not useful to reach the goal. For example, it finds out that a command for changing the brightness of a smart bulb can also be used to switch the light off.
Among the different RL algorithms tested, Q-learning presents the best performance. With tuned parameters, it learns the optimal sequence of commands after few hundreds interactions, exploring the state space of the smart bulb, which in turn has millions of states.

\TOOL demonstrates how RL solutions can be successfully exploited to support semantic interoperability, opening to possible automated solutions to discover the semantics of poorly documented IoT systems. \TOOL is open source and freely available to the community.\footnote{\url{https://github.com/SmartData-Polito/RL-IoT}}

In the remaining of the paper, after a discussion of related work in Section~\ref{sec:related}, we present the design of our \TOOL framework in Section~\ref{sec:methodology}.
Next, in Section~\ref{sec:results} we compare the performance of different algorithms, perform parameter tuning and present thorough experimental results. Section \ref{sec:conclusions} concludes our work, presenting possible future steps.

%% file: 03_related.tex
\section{Related Work}
\label{sec:related}

The work most similar to ours is~\cite{RLIoTinteroperability} where the authors propose the use of the Q-learning algorithm to facilitate the interoperability of IoT systems. However, the authors only discuss the applicability of the RL-approach to a REST-based protocol, without introducing a general system or validating the approach. Here we demonstrate the potentiality of the idea without assuming a specific protocol. We also demonstrate the feasibility of \TOOL in practice and contribute the software to the community. 

Considering the use of RL for learning protocols, most previous work targets security applications, such as honeypots. Authors of~\cite{RASSH} develop a honeypot capable of learning commands from direct interaction with attackers. Their self-adaptive honeypot emulates a SSH server and uses the SARSA RL algorithm to interact with attackers. Later, the same authors propose an improved version based on Deep Q-learning~\cite{QRASSH}. The authors of~\cite{EngagementSMDP} design another adaptive honeypot, modelling the attacker as a Semi-Markov Decision Process (SMDP) and applying RL to learn the optimal policy. 

The authors of~\cite{luo2017iotcandyjar} present adaptive honeypots for studying the security of IoT devices. They propose to use RL to automatically obtain knowledge about the behaviour of attackers, building an ``intelligent-interaction'' honeypot that could engage attackers.
Authors of~\cite{AttackDetectionRL} study IoT attacks too. The authors argue that the diversity of protocols, software and hardware of IoT devices, together with dynamic changes in attacking strategies calls for automatic ways to recognize the attacks. They use RL techniques to search for the best way to answer attackers' commands. 

All these efforts share the RL-based approach with our \TOOL framework. We however target the interoperability scenario. In contrast to security applications in which attackers often try to misuse devices and protocols, our goal is to learn how to legitimately interact with IoT devices that may be poorly documented. 

%% file: 04_methodology.tex
\section{Methodology}
\label{sec:methodology}

This section describes our methodology. We first summarize the reinforcement learning approach and algorithms in Section~\ref{sec:rl}. Then we introduce \TOOL, our framework for learning, in Section~\ref{sec:framework}, and we describe the environment in Section~\ref{sec:states}. In Section~\ref{sec:yeelight} we describe the application to the Yeelight protocol, and in Section~\ref{sec:paths} we define the goals for the smart bulb. 

\subsection{Reinforcement learning algorithms}
\label{sec:rl}

Reinforcement learning is a technique to train a system where learning is achieved by interacting with the environment. It is based on rewards and punishments~\cite{SuttonBarto}.

Formally, an agent is in a state $s \in S$ defined in function of the environment. The agent may change state following an action $a \in A$ taken at \emph{discrete} time steps. At time $t$, the agent decides which action $a_t$ to take given its current state $s_t$ and, as a consequence, it moves to $s_{t+1}$. The action then causes a change to the system state and the agent possibly receives a reward $r_{t+1}$. 

Considering the above setup, a policy $\pi$ determines the action $a$ taken by the agent when in a particular state $s$. The task of a RL algorithm is thus to determine a policy that maximises a function of the received reward. There exist several methods to search for optimal policies. Here we consider well-established algorithms that operate based on a value function $V(s)$, that represents the expected accumulated reward when starting from the particular state $s$ to follow a policy $\pi$. We include algorithms belonging to two categories:
\begin{itemize}
    \item \textit{Temporal-Difference (TD) learning}: 
    The agent updates $V(s)$ after every time step as:
    \begin{equation}
    (s_t) \gets V(s_t) + \alpha[r_{t+1} + \gamma V(s_{t+1}) - V(s_t)]\label{eq:1}
    \end{equation}
    The parameter $\alpha$ is the learning rate -- \ie how much $V(s_t)$ should change when updated. $\gamma$ is a discount factor that weights the importance of the destination state $V(s_{t+1})$. SARSA and Q-learning are popular TD algorithms~\cite{SuttonBarto}. The former is an on-policy algorithm (\ie the agent evaluates and improves only the policy $\pi$), whereas the latter is an off-policy method (\ie the agent evaluates other policies $\pi$ taking the maximum observed $V(s_{t+1})$ for updating $V(s_t)$). 
    \item \textit{TD($\lambda$) learning}: The agent takes $n$ time steps before updating $V(s_t)$. As such, TD($\lambda$) algorithms must memorize visited states to update them later. The parameter $\lambda$ controls how the $n$ future states influence $V(s_t)$ (\ie like a decay parameter). The most common TD($\lambda$) algorithms are direct extensions of traditional TD learning methods: SARSA($\lambda$) and Q($\lambda$).\footnote{Two different Q($\lambda$) versions exist: Watkin's Q($\lambda$) and Peng's Q($\lambda$) \cite{SuttonBarto}. In this work we use Watkin's version.}
\end{itemize}

Both TD and TD($\lambda$) algorithms need a strategy to select the current policy. This strategy should allow continuous exploration of new actions. The most used strategy is called $\epsilon$-greedy policy selection. The  $\epsilon$-greedy strategy balances the trade-off between exploitation (the agent selects already tried actions found to be effective in producing reward) and exploration (the agent randomly selects actions in the search for better paths). With a probability of $\epsilon$, a random action is selected. The greedy policy is instead chosen with a probability of $1-\epsilon$, selecting the action that currently has the highest value for the state $s$ inside the value function $V(s)$. $\epsilon$ can decrease over time, allowing a high exploration during the initial search. 

\subsection{IoT reinforcement learning framework}
\label{sec:framework}

\begin{figure}[t]
\centering
\includegraphics[width=.95\linewidth]{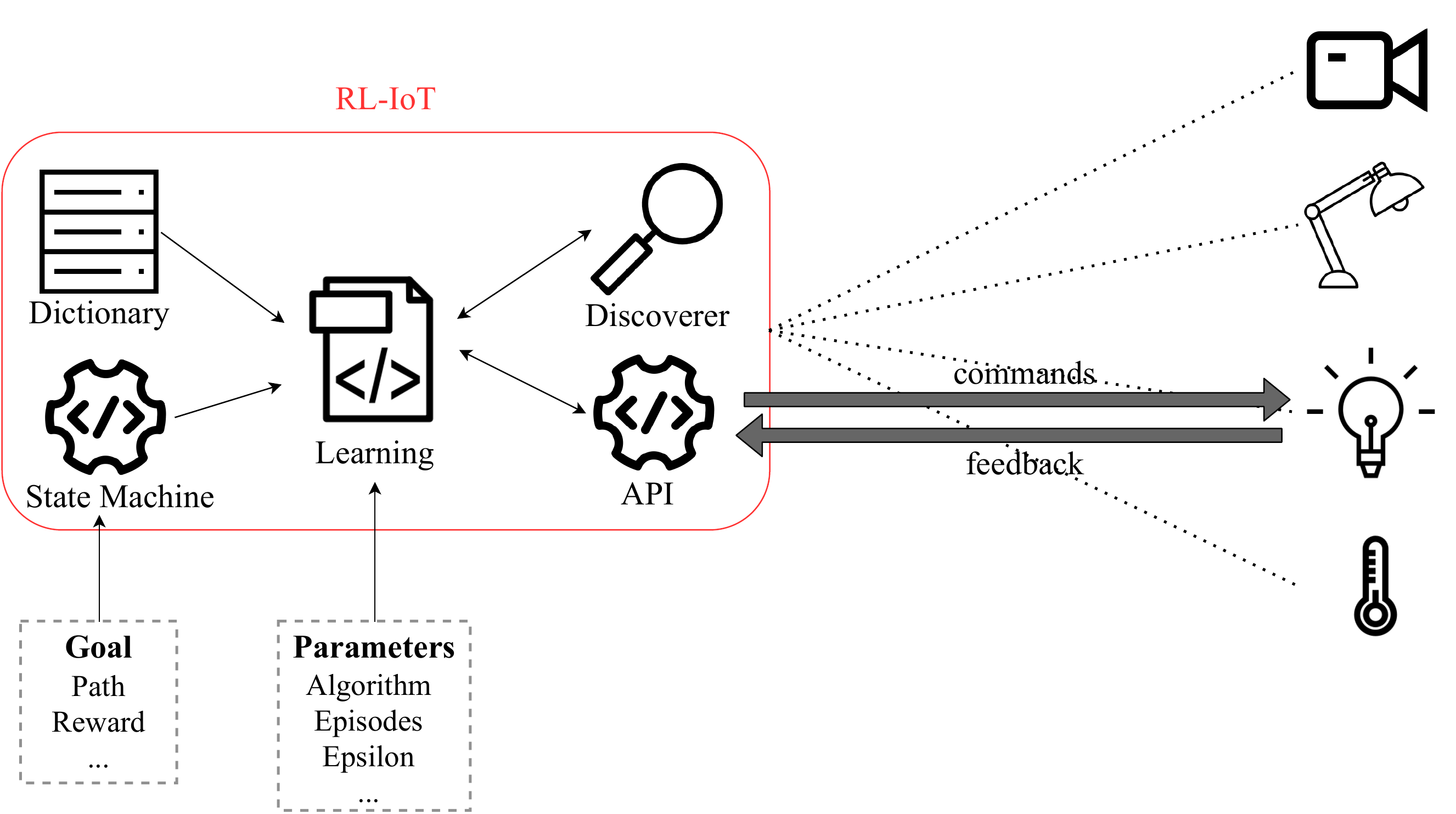}
\caption{\TOOL framework overview.}
\label{fig:framework}
\end{figure}

Figure~\ref{fig:framework} summarizes the core \TOOL framework. It receives as input a \emph{Goal} that the RL Module should learn how to achieve. The \emph{Goal} represents a sequence of settings the device should follow, i.e., paths on the device state-machine. 
This goal is device-specific, and we will detail it when discussing our case study with the Yeelight smart bulb.

\TOOL leverages an internal \emph{Message Dictionary} containing a list of IoT protocol messages that can be used to interact with devices. This dictionary can be built from protocol specifications, via automatic reverse engineering solutions or by traffic sniffing.
It can contain a mix of messages from different IoT protocols, vendors, versions, etc.

\TOOL employs state-of-the-art RL algorithms, where the \emph{Learning} module builds and updates the internal \emph{State Machine}. The Learning module supports the previously cited RL algorithms -- Q-Learning, Q-Learning($\lambda)$ SARSA and SARSA($\lambda$)~\cite{SuttonBarto}, each with its parameters. It explores which of the several messages in the Dictionary can be used to change the state of the IoT device towards the given Goal. RL algorithms exploit a reward function (custom to each path) to evaluate the benefits of each action taken by the learner in a given state. 

The Learning module interacts with two other modules. The \emph{Discoverer} module is responsible for scanning the local network in the search for IoT devices. It employs classic scanning approaches (\eg \texttt{nmap}\footnote{\url{https://nmap.org/}}) for searching online devices and performing an initial fingerprint to determine open ports. At last, the \emph{Socket API} module abstracts all the mechanisms to communicate with the target IoT device. 
Beside sending commands, it may also support the reception of feedback directly obtained from the IoT device, if available. For instance, it can support parsing messages that return the device state.

\subsection{Environment definition}
\label{sec:states}


In general, the state of a device can be represented as the powerset of all the current properties of the device, which describes its behaviour and settings -- \eg whether it is on/off and the combination of all the values of its configurable parameters.
We define the state-machine of a protocol as a graph containing nodes for states and edges for commands that let the device move from one state to another. A collection of ordered states linked by commands is a \emph{path}. Commands stored in the Message Dictionary could change the IoT device settings, \ie the current state. 
With states and commands we can define a state-action value function for the RL algorithms, described by the value-function matrix $Q$. 

The reward associated with the state-machine and the desired path can be provided to \TOOL as input, and it is used by the RL agent at each time step. 
\TOOL runs this procedure many times, i.e., for many episodes. An episode ends when the RL agent reaches the terminal state(s), or after a maximum number of iterations. 
During each step in an episode, \TOOL accumulates reward.
With such reward, the RL agent updates the state-action matrix $Q$ according to Equation~\ref{eq:1}, and uses it to select which next command to send, trying to maximize the total reward.

\subsection{Case study: The Yeelight bulb}
\label{sec:yeelight}

We use a Yeelight smart bulb as a case study to demonstrate the feasibility of our approach.\footnote{For all experiments, we use Yeelight LED Smart Bulb 1S Color (8.5W-E27-YLDP13YL) devices.}
We select this device because Yeelight provides generic protocol documentation valid for all their IoT devices.\footnote{\url{https://www.yeelight.com/download/Yeelight_Inter-Operation_Spec.pdf}}
Knowing the protocol allows us to understand and validate what \TOOL can learn.
The protocol offers 37 commands, and only about half of them work with the selected smart bulb, with multiple commands that could generate the same action. For instance, one could set a color via a $set\_rgb$, $set\_scene$, or $adjust\_prop$ message. 

Yeelight devices connect to the network using Wi-Fi. After the initial setup, the device periodically broadcasts its presence using \emph{advertisement} UDP messages. It is thus easy for the Discoverer module to find them in the LAN.
Once \TOOL identifies the device IP address, it starts interacting with it sending messages from the Dictionary. Yeelight offers control protocols running on top of both HTTP and raw TCP sockets. The latter relies on simplistic \emph{JSON} messages that carry commands. Figure~\ref{fig:yeelight} presents one of the simplest \emph{JSON} messages to set the color of a smart bulb.
The device responds to well-formatted commands with a \emph{result} message - on the bottom of Figure~\ref{fig:yeelight}. Other commands allow clients to obtain information about the state of the device, to change its name, to turn light on and off, to change light intensity, to play music, to set fan speed, etc. As said, not all commands are supported by our smart bulb.

\begin{figure}[h]
\small
Command: 
\begin{verbatim}
{"id": 1, 
 "method": "set_rgb", 
 "params": [255, "sudden", 0]}\r\n
\end{verbatim}
Answer: 
\begin{verbatim}
{"id": 1, "result": ["ok"]}\r\n
\end{verbatim}
\caption{Examples of Yeelight protocol messages.}
\label{fig:yeelight}
\end{figure}

The  commands can have some parameters to set.
While these parameters usually belong to finite sets, for some commands the number of admissible values can be huge (like for integer or string parameters).
Indeed the combinations of commands and their parameters result into more than $10^9$ distinct combinations that the RL agent could send to a Yeelight device. 
For our case study we simplify the definition of our environment 
according to our goal. 
To reduce the action space, we consider the action as only one command, with its parameters that we randomly choose in valid ranges.

\subsection{Case study: Definition of goals}
\label{sec:paths}

For testing \TOOL, we build and study two scenarios with different state-machines of increasing complexity.

\begin{figure}[t]
\centering
\includegraphics[width=.67\linewidth]{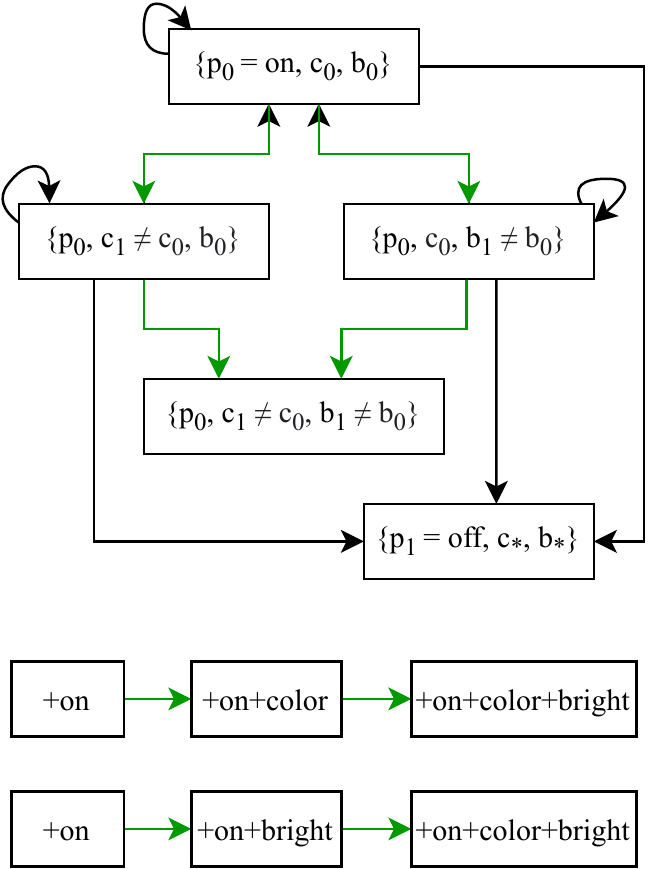}
\caption{Goal 1: Simple state-machine where we want to learn how to change the color and the brightness of the bulb, in whatever order. The ``*'' refers to whatever value.}
\label{fig:path1}
\end{figure}

In the first scenario, given a switched-on bulb, our Goal~1 is to learn how to change the color and the brightness of the bulb, in whatever order. In Figure~\ref{fig:path1} we report the state-machine for this first scenario. Each state considers different attribute values: power $p$, color $c$ and brightness $b$. Hence the state is defined by the values of the tuple $\{p,c,b\}$.  
We disregard the other attributes of the light configuration.
Here, we have two final states, where an episode will successfully end: either we reach our goal  ($\{p_0=\text{on}, c_1\ne c_0, b_1\ne b_0\}$) or we fail, i.e., we turn off the bulb too early without setting color and/or brightness ($\{p_1=\text{off}, c_*, b_*\}$).


We perform a transition from one state to another inside the state-machine when a command modifies one or more of these attributes.
With this strategy for defining state-machines, we are able to condensate multiple states into a single one as in the figure. This allows us to represent cases in which an attribute is continuous and/or has a high number of admissible values, i.e., the color of the smart bulb has 16777216 possible values.
In Figure~\ref{fig:path1} we draw possible transitions (arrows) only if a command exists in the protocol to change such property.
The actions (commands and their parameters) are not specified in the picture since there might be multiple commands that could produce the same transition. 
Similarly, there exist a lot of commands that do not change the state, represented as state self-transition (a looped arrow).
Note that it is even possible to get back to a previous state (e.g., setting back the original color $c_0$).

\begin{figure}[t]
\centering
\includegraphics[width=.98\linewidth]{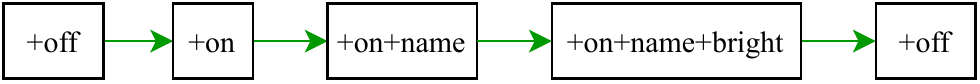}
\caption{Optimal path for Goal 2.}
\label{fig:path2}
\end{figure}

\begin{figure*}[t]
\centering
\includegraphics[width=.98\linewidth]{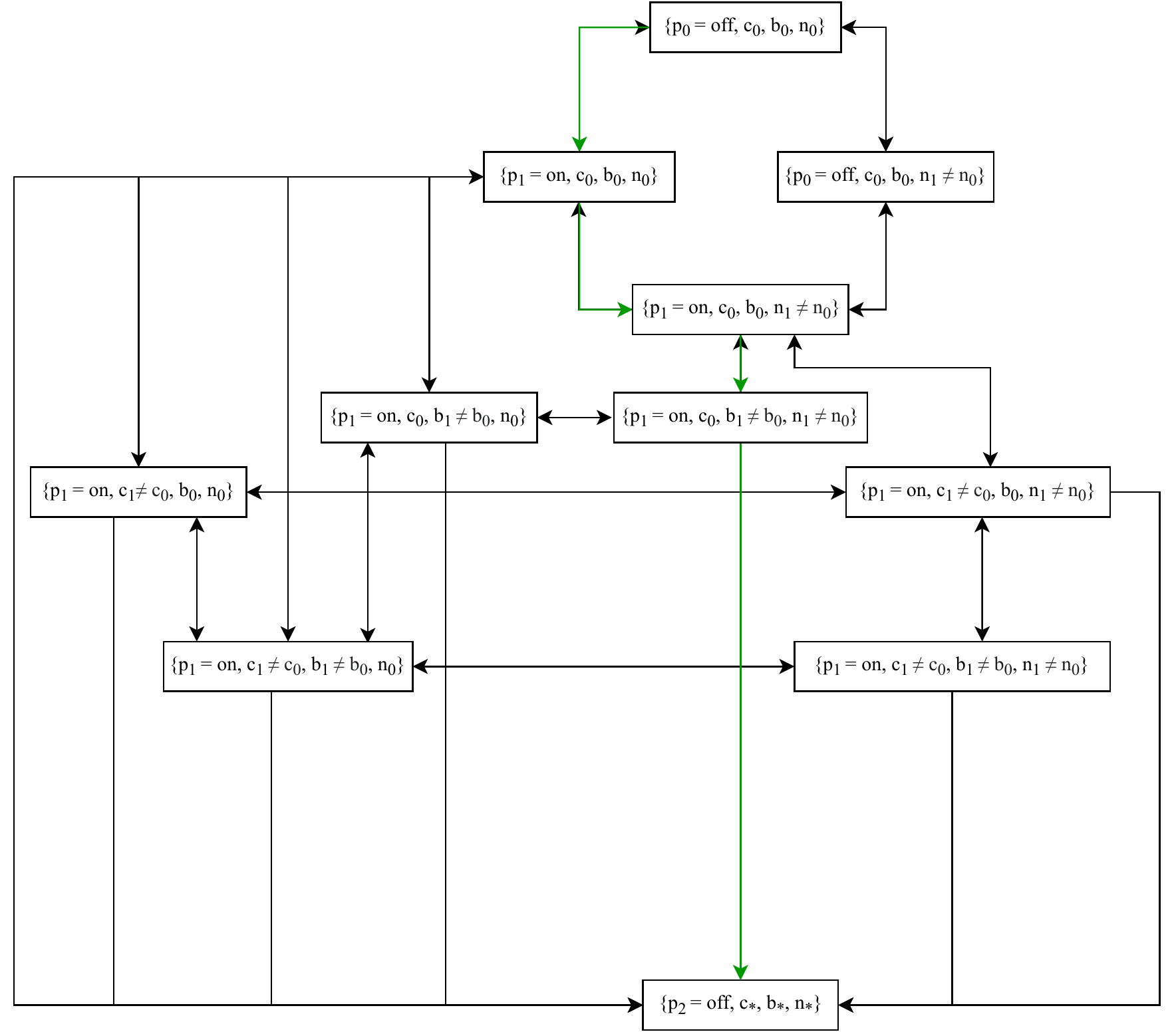}
\caption{Goal 2: Complex state-machine where we want to learn how to turn on the bulb, change its name, its brightness and finally turn it off, in this specific order and without changing the color.}
\label{fig:complete-path2}
\end{figure*}

On the bottom part of Figure~\ref{fig:path1} we show the optimal paths, i.e., the shortest sequences of state changes we want to learn: the name assigned to each box refers to the total modified attributes so far. The optimal policy for Goal~1 visits 3 states with 2 actions, i.e., requiring 2 time steps the least.
Here, we assign the rewards as follows: (i)~each new issued command has a small additional negative reward ($-1$), since we want to reach the goal in as few steps as possible; (ii)~we give higher negative reward ($-10$) when the command produces an error and the state does not change; (iii)~we assign no reward when we reach the final state without completing the path $\{p_1=\text{off},c_*,b_*\}$; (iv)~we give large positive reward ($+205$) when we reach the desired final state $\{p_0,c_1\ne c_0,b_1\ne b_0\}$. 
Hence, with these assigned rewards, the optimal paths will reach a total reward of 203 (i.e., 205 minus 2 steps). 


With similar considerations, we draw and implement also another state-machine that we call Goal~2, shown in Figure~\ref{fig:complete-path2}. The optimal path can be found in Figure~\ref{fig:path2}.
The specific goal we want to learn is, in this specific order: (i) turn the bulb on, (ii) change the device name, (iii) change brightness, and (iv) turn the bulb off.
Here our goal is more complex since we want to learn how to move through a specific sequence of states. 
Since we add the $name$ attribute among those we want to change, the state definition becomes \textit{\{power, color, brightness, name\}}.
We also require the bulb color to remain constant, and thus the color is still considered as part of the state definition. 

We assign a large positive reward (+222) at the final state if we pass through the desired states in Figure~\ref{fig:path2} in the right order.
If we arrive to the same final state, but in a different sequence of the same intermediate states, we assign a positive, but smaller reward (+200). Negative rewards are similar to Goal~1.
Here the optimal path is unique, with an optimal length of 4 time steps, generating the maximum total reward of 218 (i.e., 222 minus 4 steps).

\subsection{Performance metrics}

\begin{table}[t]
\caption{Formal notation for evaluation metrics and parameters of the RL algorithms.}
\begin{tabular}{lll}
$E$ & & episode \\
$N_E$ & & total number of episodes \\
$R(E)$ & & total reward obtained in episode $E$ \\
$T(E)$ & & total number of time steps $t$ in episode $E$ \\
$N_a$ & & total number of actions performed \\
$C(n_a)$ & & cumulative reward obtained after $n_a$ actions \\
$Q(s,a)$ & & action value function or Q value function \\
$\epsilon$ & & exploration-exploitation trade-off \\
$\alpha$ & & learning rate \\
$\gamma$ & & discount factor \\
$\lambda$ & & trace decay \\
\end{tabular}
\label{tab:notation}
\end{table}

We consider three metrics for evaluating results and comparing the performance of the various algorithms.

We summarize the notation we use in Table \ref{tab:notation}.
Note that we assume that the sets of states $S$ and actions $A$ are finite sets.
If not, there exist methods which combine standard RL algorithms with function approximation techniques, such as neural networks~\cite{DQN,ContinuousStateActionSpaces}.
Having finite sets the Q value function $Q(s,a)$ can be represented as a matrix.

In our scenarios, a terminal state always exists.
We call this $T(E)$, i.e., the number of time steps used in a single episode $E$ to reach the terminal state. We force $T(E)<T_{\max},\ T_{\max}=100$.
We compute the total reward $R(E)$ obtained during episode~$E$:
$$R(E) = \sum_{t=1}^{T(E)} r_t (E)  \text{ for }  \: E \in \{1,..., N_E\},$$ 
being $N_E$ the total number of episodes we let \TOOL run. 

These metrics can be averaged over multiple executions - which we call \textit{runs} - of the learning process.  
Similarly, we compute the moving average for a specified window size~$w$. Average and moving average help to appreciate the learning curve which is affected by the randomness present in each run due to exploration. 

Finally, we compute the cumulative reward $C(n_a)$ from the beginning of the learning process over the number of actions performed~$n_a$:

$$C(n_a) =  \sum_{E=1}^{E_{n_a}} \sum_{t=1}^{T_{n_a}(E)} r_t(E) \text{ for }  \: n_a \in \{1,..., N_a\}$$

This metric takes into account not only the reward reached within an episode $E$, but also how much reward cumulatively was obtained until that episode.
Indeed, the number of episodes to consider $E_{n_a}$ depends on number of actions $n_a$. Also the number of time steps to consider $T_{n_a}(E)$ depends on the episode: it is $T(E)$ if $E\ne E_{n_a}$;  or the number of remaining time steps to reach the limit of total number of actions imposed by $n_a$ in the last episode $E=E_{n_a}$.
To compare different algorithms, we compute the average among different runs, as for $T(E)$ and $R(E)$. 
Here, the difference is that we ``consume'' the same number of actions $n_a$ after a different number of episodes $E_{n_a}$ in different runs. 

%% file: 05_results.tex
\section{Results}
\label{sec:results}

In this section we summarize the results. In Section~\ref{sec:result_paths} we evaluate whether \TOOL can learn the given target paths. 
In Section~\ref{sec:result_parameters} compare different RL algorithms with tuned parameters. Finally, Section~\ref{sec:result_costs} studies the cost of training the RL models in terms of training time and network traffic.  
For all experiments, \TOOL runs on a x86-64 PC with 4GB of RAM and two cores, connected to the same Wi-Fi network as the Yeelight bulb.

\subsection{Learning capability}
\label{sec:result_paths}

We start focusing on whether \TOOL can learn how to reach the desired goals. We apply the Q-learning algorithm while observing the reward evolution over episodes, and the number of time steps needed to arrive to the target state at each episode.
In order to provide an intuition of how \TOOL interacts with the smart bulb while exploring possible commands, we share a video of one run at \url{https://tinyurl.com/yws6m7ec}.

\begin{figure}[t]
  \centerline{\includegraphics[clip=true, width=0.24\textwidth]{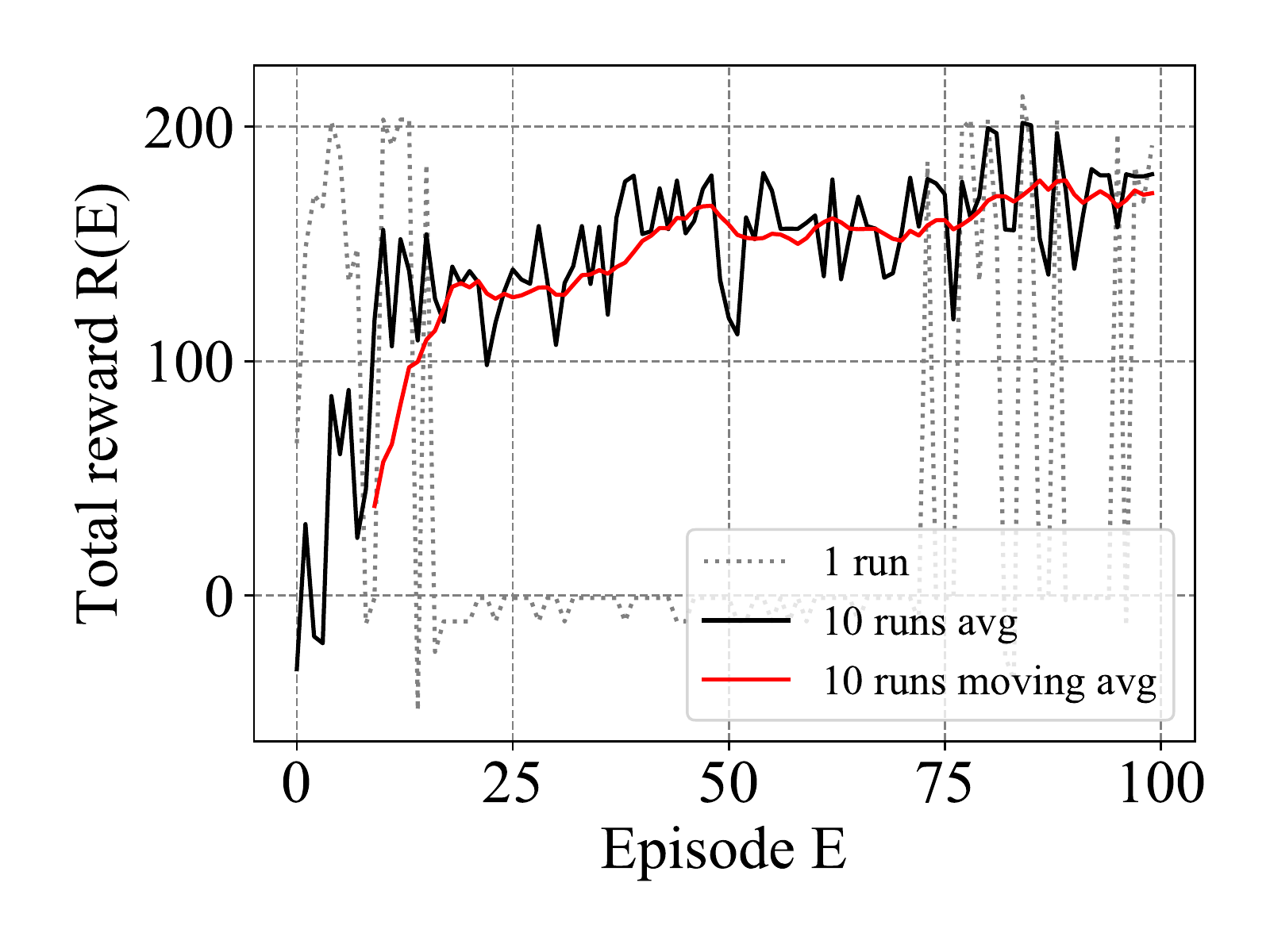}
  \includegraphics[clip=true,  width=0.24\textwidth]{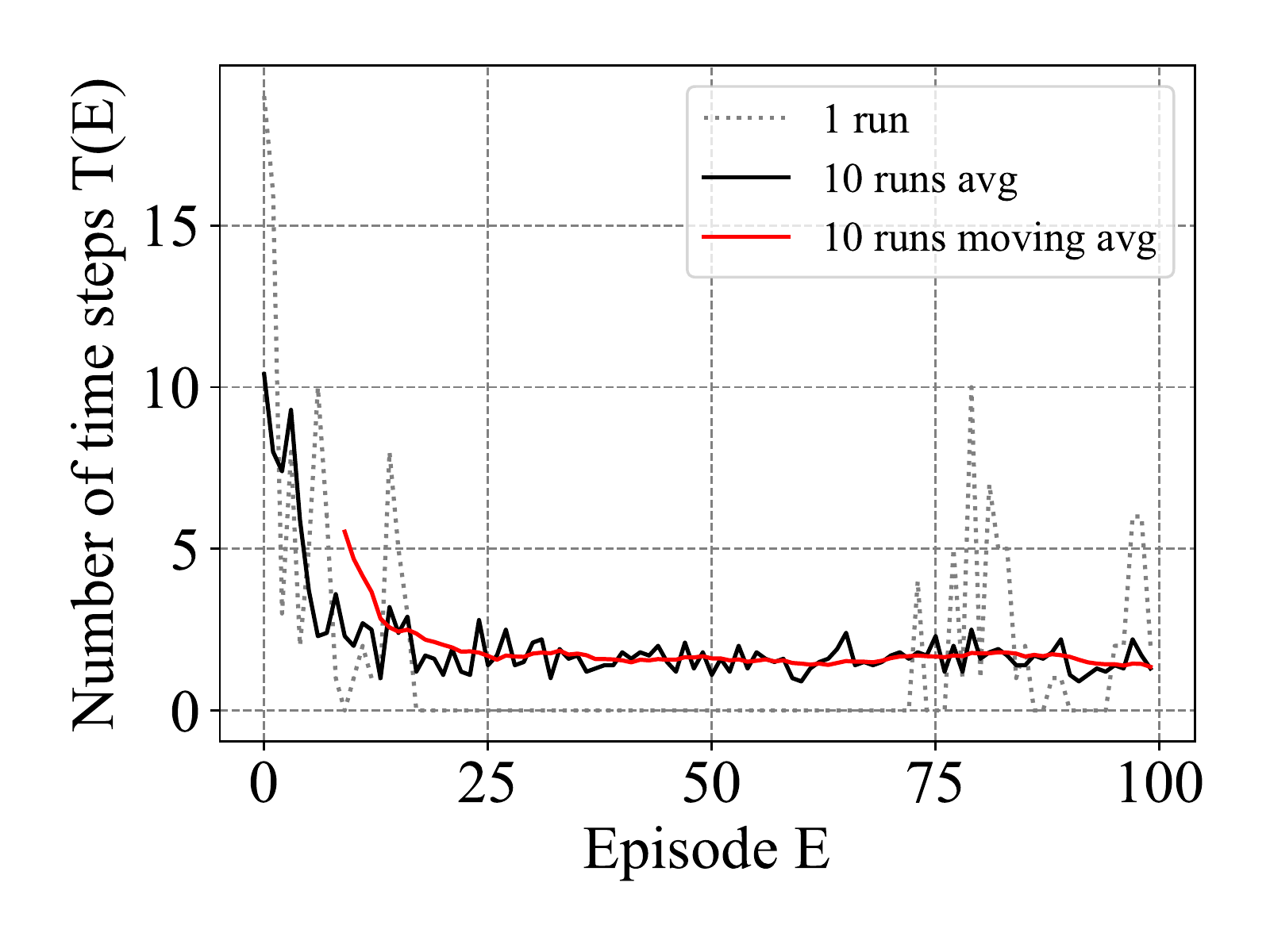}}
  \caption{Q-learning performance while learning Goal~1. $\epsilon=0.2$, $\alpha=0.1$, $\gamma=0.55$.}
  \label{fig:qlearning-trivial-path}
\end{figure}

\begin{figure}[t]
  \centerline{\includegraphics[clip=true, width=0.24\textwidth]{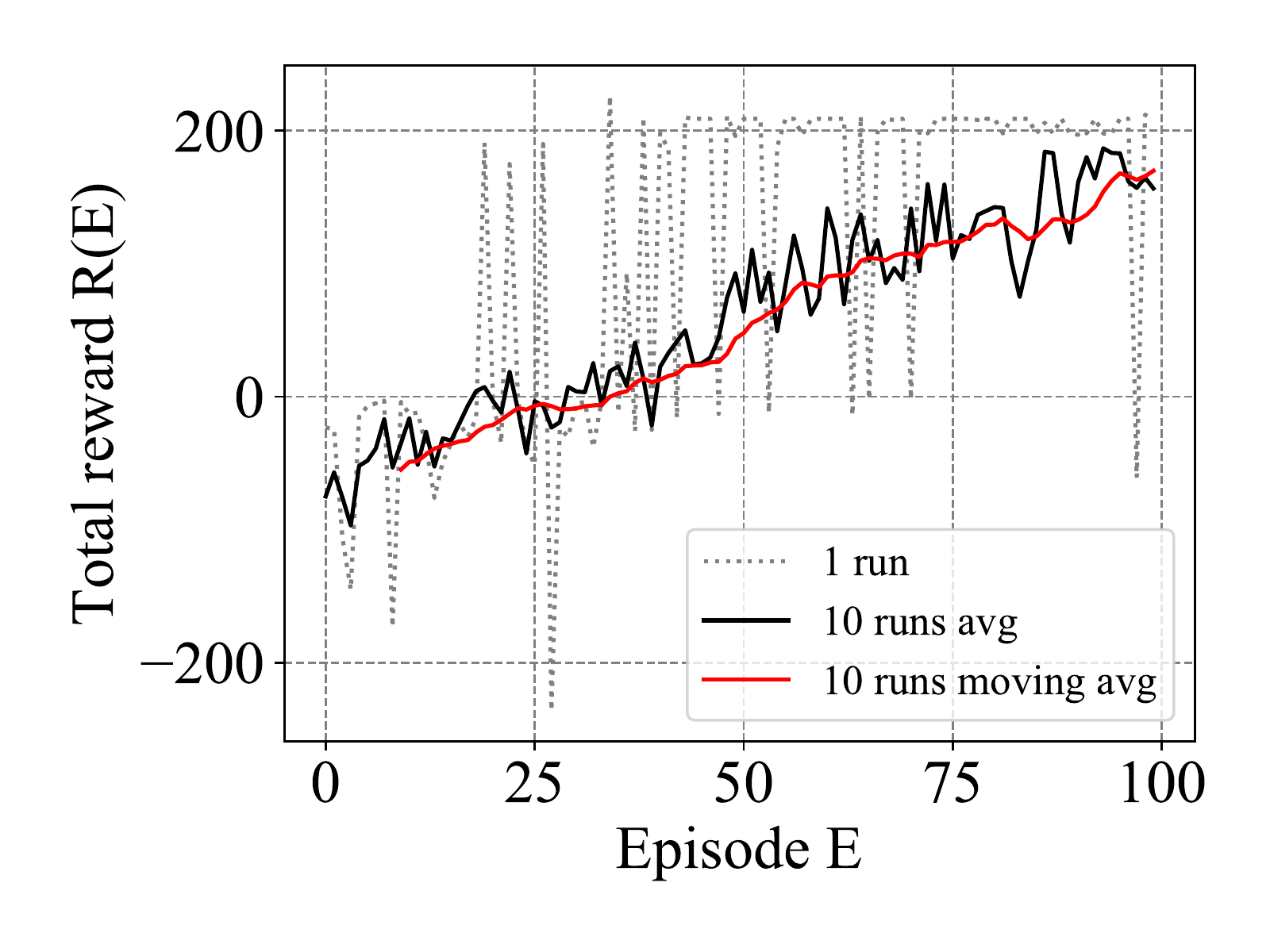}
  \includegraphics[clip=true,  width=0.24\textwidth]{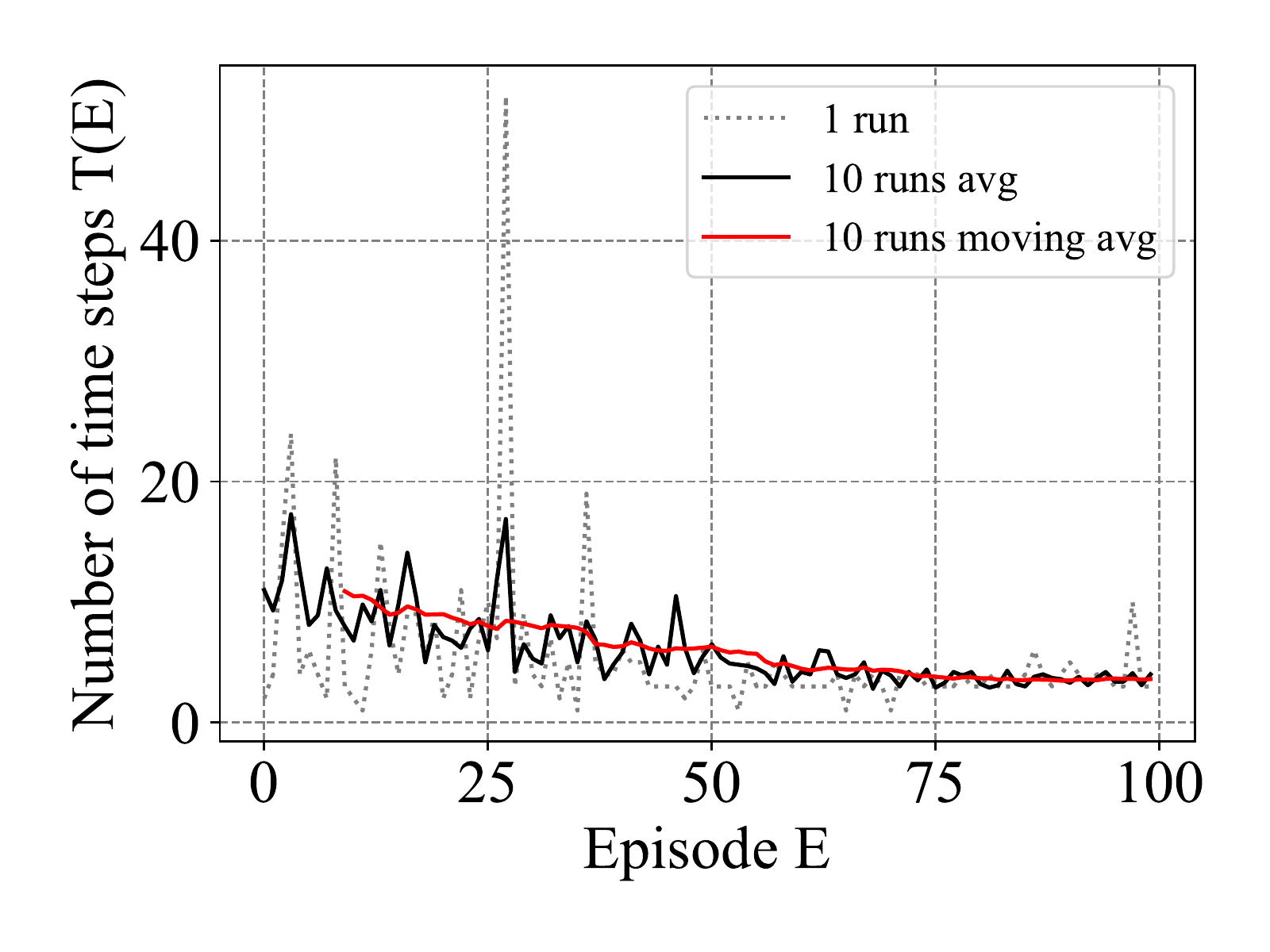}}
  \caption{Q-learning performance while learning Goal~2. $\epsilon=0.2$, $\alpha=0.1$, $\gamma=0.55$.}
  \label{fig:qlearning-path2}
\end{figure}

\begin{figure*}[t]
\centering
\includegraphics[width=.9\linewidth]{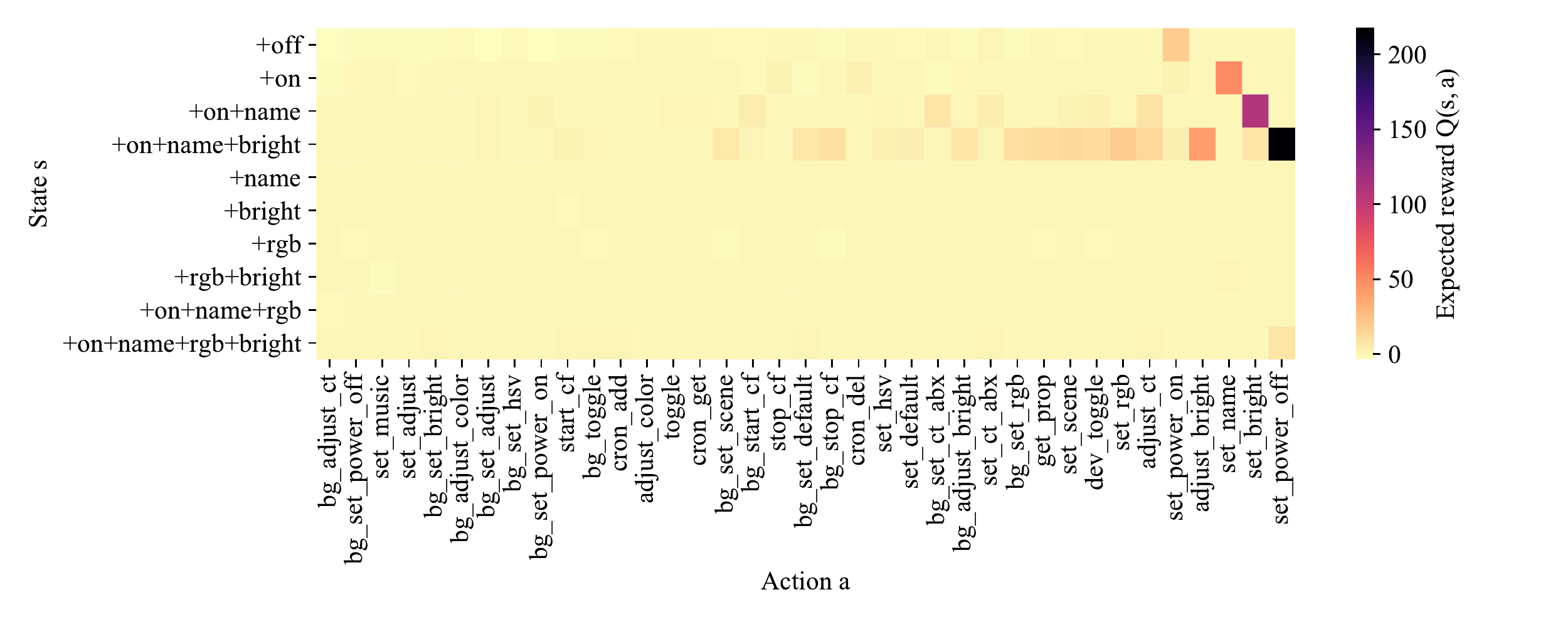}
\caption{Example of the final action-value Q matrix for Goal 2. Darker colors show commands (columns) that result in higher expected rewards for the  states (rows). $\epsilon=0.2$, $\alpha=0.1$, $\gamma=0.55$.}
\label{fig:heatmap}
\end{figure*}

Figure~\ref{fig:qlearning-trivial-path} reports the total reward $R(E)$ (left plot) and the number of time steps $T(E)$ (right plot) of each learning episode for Goal~1.
Dotted gray line details a single Q-learning run; solid black line reports the average of 10 runs; red line shows the moving average over the 10-run, taking into account a window $w$ of 10 episodes. 

Q-learning initially cannot reach the desired state.
Missing the large positive rewards, it accumulates a negative reward on average. After few episodes, $R(E)$ grows to the maximum value that could be observed (203 here). However, comparing the line for a single run to the average over 10 runs we observe a lot of variability. This can be explained by the random exploration component (controlled by $\epsilon$) in the Q-learning algorithm. This exploration phase may penalise the single episode with low final reward, even if the system has already discovered the target goal before.
The right plot in Figure~\ref{fig:qlearning-trivial-path} shows that Q-learning finds how to reach the desired state with very few actions. After around 15 training episodes, on average, it finds policies composed by 2 or 3 steps, thus the average reward gets closer to the maximum. Recalling that for the trivial Goal~1 scenario the optimal path is composed by 2 steps, we conclude that Q-learning has already found the best path to the goal after 15--20 training episodes.

We report the same results for Goal~2 in Figure~\ref{fig:qlearning-path2}.
As before, we depict only results for Q-learning, and lines show numbers for a single run, 10-run average and moving average. 
The results are qualitatively similar for Goal~2, but with slower learning, given the higher complexity of the goal.
However, also in this case the learning phase is still able to discover paths with positive reward after around 20 episodes. Given the large state space to explore, the algorithm is still improving its performance even after 100 episodes. The number of steps (right plot) is often below 7 even after few episodes, meaning that (on average) the algorithm is moving around the optimal path (with 4 steps). 

To give the intuition of the learning process achieved by \TOOL, we depict in Figure~\ref{fig:heatmap} the $Q$ matrix for Goal~2 obtained after 100 episodes. Rows represent states, with the first 4 rows being the desired optimal path in the second scenario (cfr. Figure~\ref{fig:path2}). Columns represent all available commands (actions). The darker is the color, the higher is the chance to select that command in that state. To ease the visualization, we sort commands by increasing reward.
Observing the cells with darker colors, we see that Q-learning has indeed learned the expected sequence of commands to follow the given goal: when off - turn on the lamp, then set the name, the brightness, and at last turn the lamp off. Interestingly, \TOOL has also identified alternative valid commands to move to the desired state, as shown by moderately darker shades for some commands.

For example, the algorithm is able to identify several ways to turn the lamp off when in the \textit{+on+name+bright} state -- besides the $set\_power\_off$ command. For instance $adjust\_bright$ to 0, or $set\_rgb$ to 0.  In other words, \TOOL discovers multiple ways to perform the same task from its interactions with the environment.

This shows the potential of \TOOL in supporting the discovery of semantics of IoT messages. With our use case we can easily verify the actual command semantics. Yet in the general case this could not be easy, e.g., when the protocol uses binary format.

\subsection{Algorithms comparison}
\label{sec:result_parameters}

\begin{figure*}[htbp]

\centering
\subfloat[Tuning $\epsilon$]{\includegraphics[width=.32\linewidth]{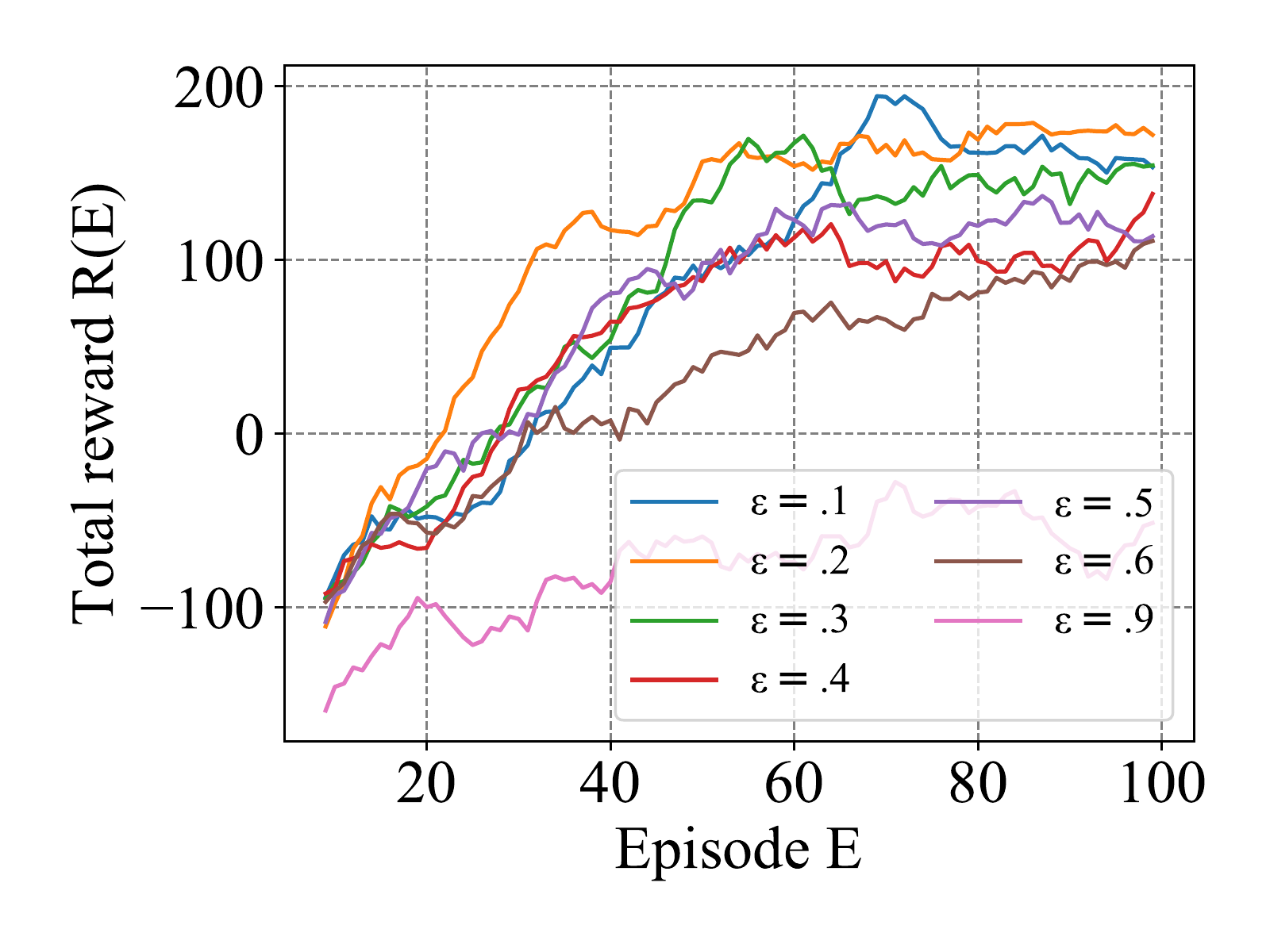}}\hfil
\subfloat[Tuning $\alpha$]{\includegraphics[width=.32\linewidth]{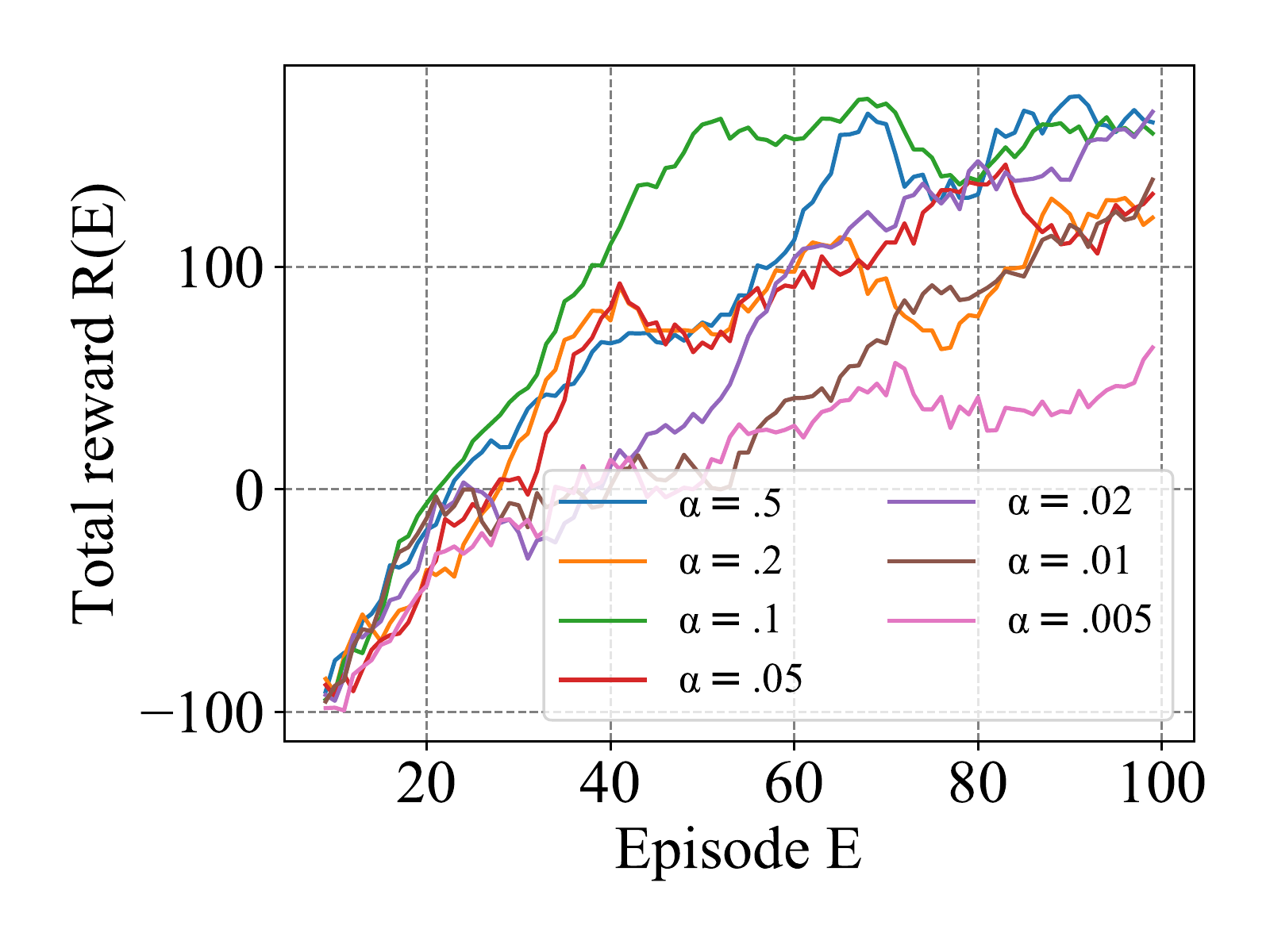}}\hfil 
\subfloat[Tuning $\gamma$]{\includegraphics[width=.32\linewidth]{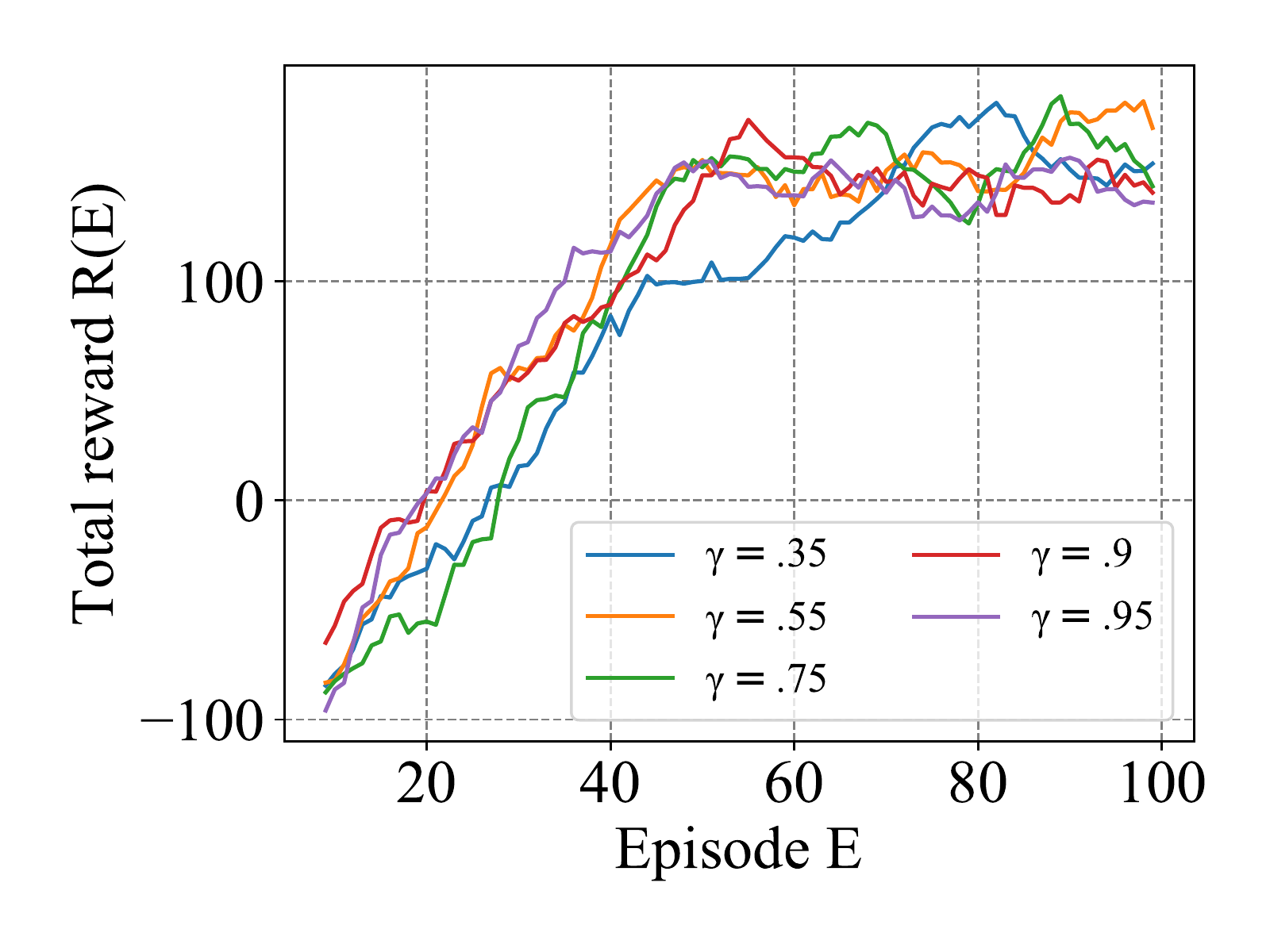}} 
\caption{Q-learning performance for Goal 2: (a) tuning of $\epsilon$  with $\alpha=0.05$, $\gamma=0.95$; (b) tuning of $\alpha$  with $\epsilon=0.2$, $\gamma=0.95$; (c) tuning of $\gamma$ with $\epsilon=0.2$, $\alpha=0.1$.}
  \label{fig:qlearning-tuning}
\end{figure*}

To compare different RL algorithms and parameter impact, we tune parameters to find the best configuration for each algorithm. 
We only report results on Goal~2, since it is more complex.

Even with the reduced commands and states,  
performing an exhaustive search for all the combinations of algorithm parameters is unfeasible. 
That is because \TOOL needs around 40 minutes to execute $100$ episodes, due to rate-limits caused by the Yeelight protocol.
We thus perform greedy experiments, in which we vary only one parameter at a time to understand its impact on results. 
More specifically, starting from values
suggested by \cite{SuttonBarto} and \cite{RLParameters} (i.e., $\epsilon=0.6$, $\alpha=0.05$ and $\gamma=0.95$), we first tune $\epsilon$ with $\alpha$ and $\gamma$ fixed. Then, we fix the best $\epsilon$, and optimize $\alpha$. Finally, we optimize $\gamma$ given the best $\epsilon$ and $\alpha$. Since TD learning algorithms are equivalent to TD($\lambda$) learning with $\lambda=0$, the best values for $\epsilon$, $\alpha$ and $\gamma$ are used with SARSA($\lambda$) and Q($\lambda$) too, with an extra optimization round for $\lambda$. We perform 5 independent runs for each parameter combination, and report the average performance. 

Figure~\ref{fig:qlearning-tuning} shows results with Q-learning changing the value of $\epsilon$, $\alpha$ and $\gamma$.
We see that $\epsilon$ (\ie the exploration-exploitation trade-off) and $\alpha$ (learning rate) are the parameters affecting the most the RL algorithm. Indeed, Figure~\ref{fig:qlearning-tuning}(a) shows that large $\epsilon$ can even prevent the algorithm to reach the maximum reward. In a nutshell, better not explore too much. Similar comment applies to low values of $\alpha$ in Figure~\ref{fig:qlearning-tuning}(b). I.e., better to learn fast. The parameters $\gamma$  in Figure~\ref{fig:qlearning-tuning}(c) and $\lambda$ (not shown for brevity) have smaller impacts on results.

After parameter tuning we obtain $\epsilon=0.2$ and $\alpha=0.1$. 
For SARSA and SARSA($\lambda$) we obtain $\gamma=0.75$, while for  Q-learning and Q($\lambda$) we get $\gamma=0.55$. 
Finally,  $\lambda=0.9$  results the best for Q($\lambda$), and $\lambda=0.5$  for SARSA($\lambda$). 
Notice that in Section~\ref{sec:result_paths}, we already used the tuned parameters here described.

\begin{figure}[tbp]
  \centerline{\includegraphics[clip=true, width=0.32\textwidth]{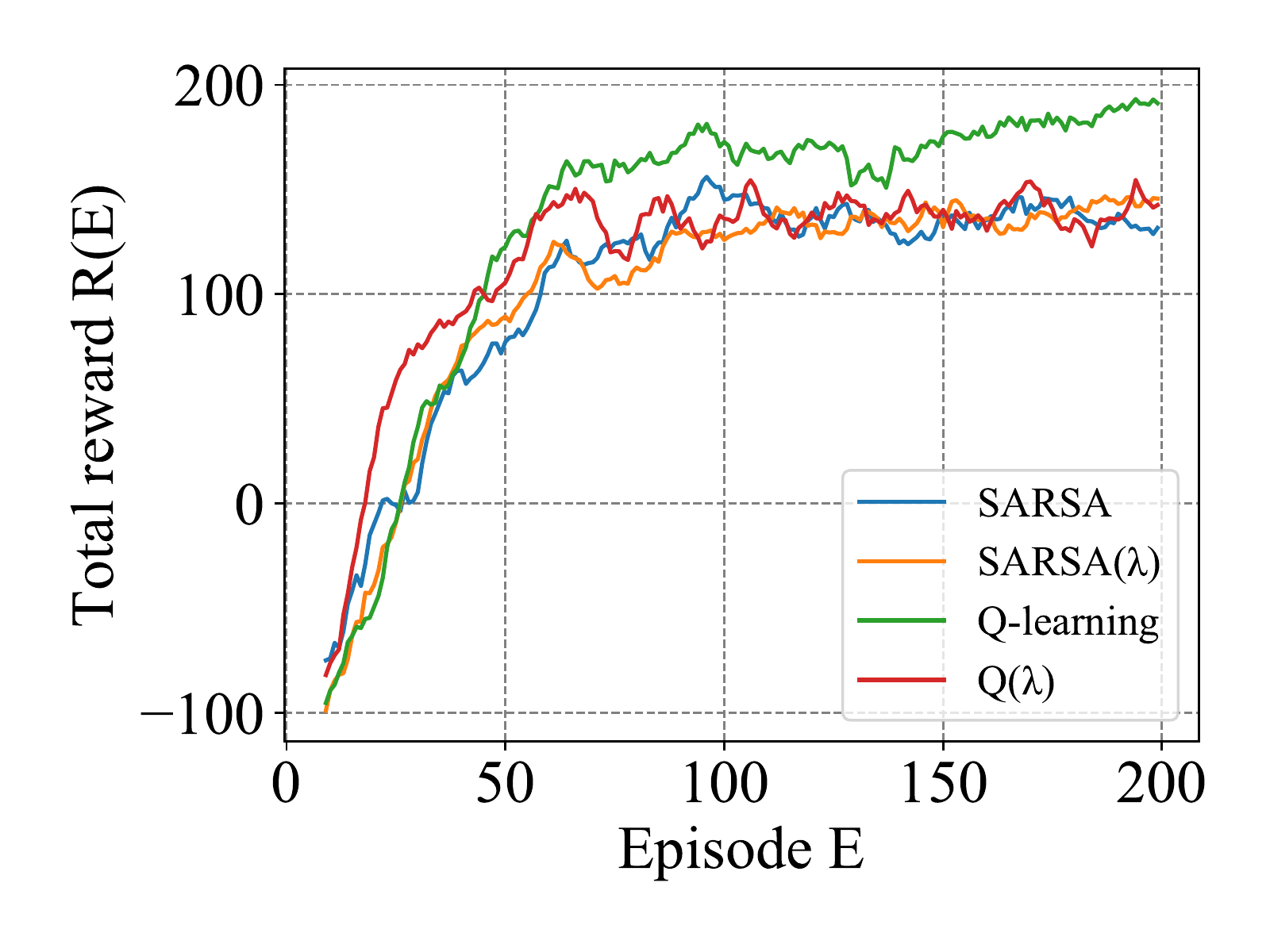}}
  \centerline{\includegraphics[clip=true,  width=0.32\textwidth]{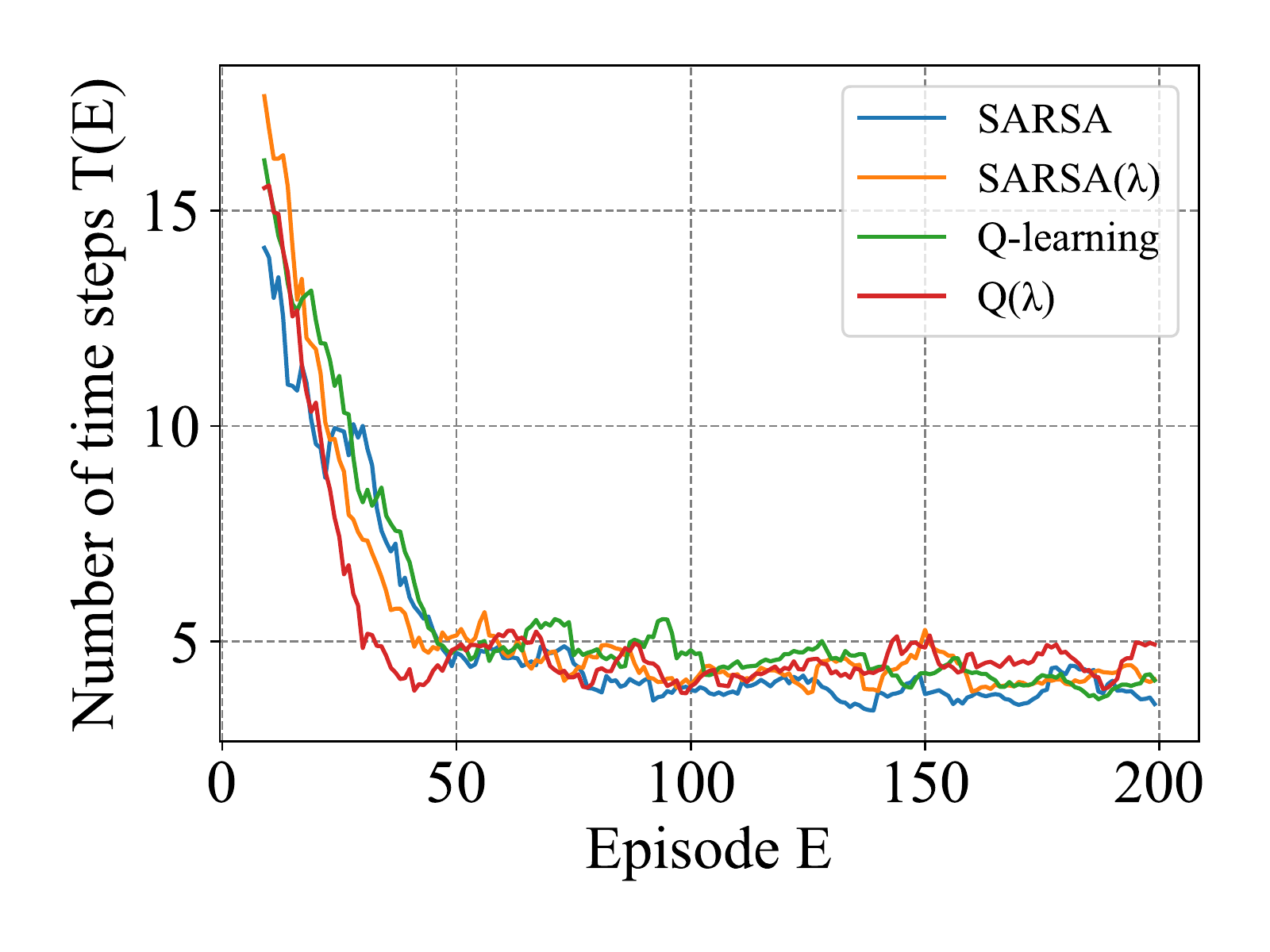}}
  \caption{Algorithm comparison for Goal~2.
  SARSA: $\epsilon=0.2$, $\alpha=0.1$, $\gamma=0.75$;
  Q-learning: $\epsilon=0.2$, $\alpha=0.1$, $\gamma=0.55$;
  SARSA($\lambda$): $\epsilon=0.2$, $\alpha=0.1$, $\gamma=0.75$, $\lambda=0.5$;
  Q($\lambda$): $\epsilon=0.2$, $\alpha=0.1$, $\gamma=0.55$, $\lambda=0.9$.
  }
  \label{fig:algos-after-tuning-path2}
\end{figure}

With these values, in Figure~\ref{fig:algos-after-tuning-path2} we compare the best configuration for the four algorithms, in Goal~2. The top plot shows the moving average ($w=10$) of the total reward $R(E)$, while the bottom plot depicts the number of time steps obtained per episode. Again, we compute the average per episode over 10 repetitions of each experiment. 

We conclude that Q($\lambda$) obtains the highest rewards during the initial 50 episodes. In other words, the algorithm learns faster than others. Yet, from episode 50 onward Q-learning wins, reaching the maximum values at around 200 episodes. We see in the bottom plot that Q-learning learns shorter paths (on average) after 200 episodes. That is, it reaches the goal with less steps, obtaining higher rewards than the other algorithms.

\subsection{Training costs}
\label{sec:result_costs}

We now evaluate the costs of training the RL algorithms in terms of number of commands sent to the IoT devices. Since 
Yeelight protocol has a rate limit on requests, we need to pace \TOOL to avoid passing these limits and triggering the device protections. 
Therefore, \TOOL  needs to minimize the number of commands to achieve satisfactory learning in real scenarios.

\begin{figure}[tbp]
  \centerline{\includegraphics[clip=true, width=0.32\textwidth]{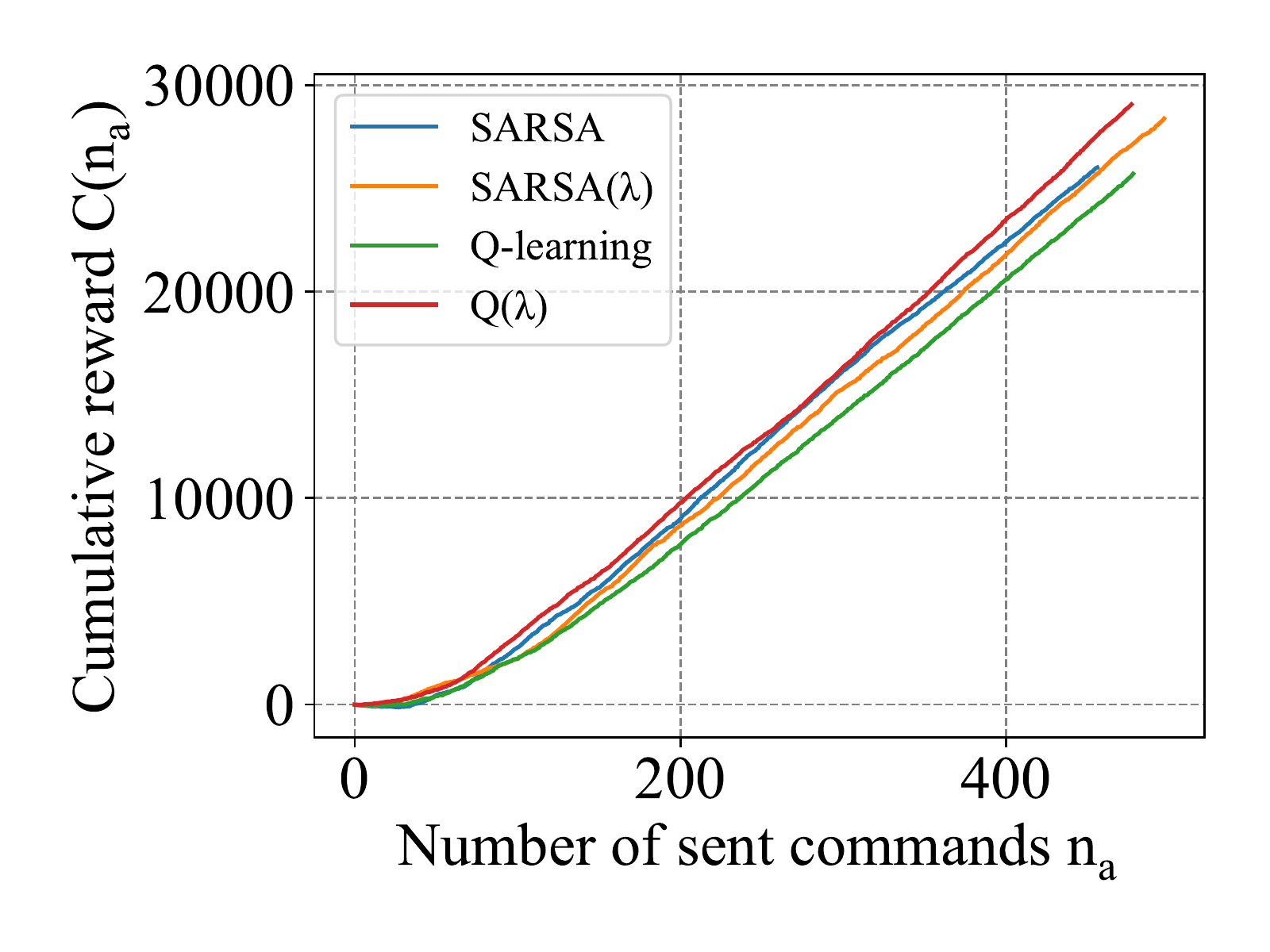}}
  \centerline{\includegraphics[clip=true,  width=0.32\textwidth]{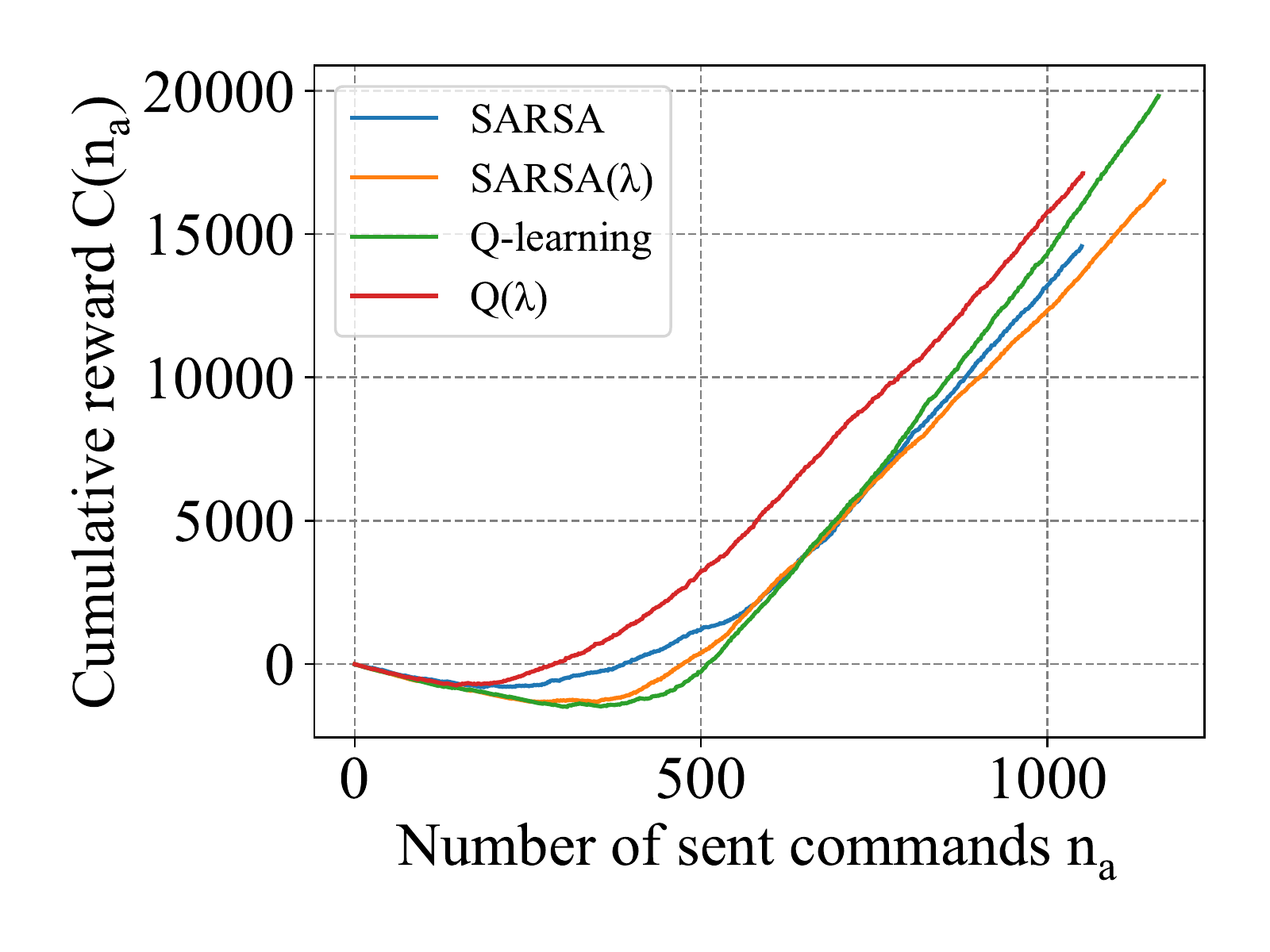}}
  \caption{Cumulative reward as a function of the number of commands for Goal~1 (top) and Goal~2 (bottom)  (200 episodes).}
  \label{fig:algos-after-tuning-path2-training-traffic}
\end{figure}

Figure~\ref{fig:algos-after-tuning-path2-training-traffic} depicts the cumulative reward $C(n_a)$ obtained by each algorithm as a function of the number of commands $n_a$ sent to the device. The plot depicts numbers for Goal~1 and Goal~2.\footnote{Different algorithms have a variable maximum number of commands $n_a$ because they might spend different number of commands to reach the end of the episodes.}

In the plot for Goal~1 (the simplest target) all algorithms behave similarly.
After exchanging around 70 commands, all algorithms start to accumulate a positive reward that since then grows linearly with the number of commands. In sum, all algorithms have learned the target path with few commands, and more training time and exploration do not result in further gains.
The plot for Goal~2 instead shows a more interesting pattern, since the complexity of the path better tests the capabilities of the RL algorithms. We see that all algorithms start with a negative accumulated reward. Some algorithms (\eg Q-learning) need to send around 400 commands before starting accumulating a positive reward. In line with results shown in Figure~\ref{fig:algos-after-tuning-path2}, Q($\lambda$) is the fastest to reach positive reward, needing around 250 commands. Whereas Q-learning is the last one to see positive numbers, its accumulated reward grows faster than others after sending around 600 commands, again confirming results seen in Figure~\ref{fig:algos-after-tuning-path2}. 

All in all, we conclude that Q($\lambda$) is able to learn solutions leading to positive reward faster for Goal~2. Standard Q-learning, while requiring more commands than others, is the algorithm able to accumulate more reward. SARSA and SARSA($\lambda$) show figures in between the alternatives.

%% file: 06_conclusions.tex
\section{Conclusions}
\label{sec:conclusions}

We proposed \TOOL, a system based on reinforcement learning that learns how to automatically interact with IoT devices. 
Given a dictionary of possible messages, the system learns which ones to send to the device to achieve a given goal. We showed the effectiveness of \TOOL in a case study with a Yeelight smart bulb. We were able to learn non-trivial patterns with as few as 400 interactions while also discovering alternative solutions. \TOOL opens the opportunity to use reinforcement learning to automatically explore the state machine of unknown protocols, thus assisting on the interoperability of IoT devices.

As future work, we will extend our experiments to make \TOOL interact with devices of multiple vendors. In this way, we will verify that \TOOL can learn the different commands to achieve a single goal on multiple devices, hopefully demonstrating interoperability in practical cases.